\documentclass{article}

\PassOptionsToPackage{numbers, compress}{natbib}

\usepackage[preprint]{neurips_2026}
\usepackage{amsmath}
\usepackage{amssymb}
\usepackage{graphicx}
\usepackage{multirow}
\usepackage{multicol}
\usepackage[table]{xcolor}

\usepackage[most]{tcolorbox}
\usepackage{fancyvrb}

\newtcolorbox{promptbox}[1]{%
    colback=gray!5,
    colframe=gray!75,
    fonttitle=\bfseries,
    title=#1,
    arc=2mm,
    boxrule=0.5pt,
    left=3mm, right=3mm, top=2mm, bottom=2mm,
    enhanced,
    breakable
}

\definecolor{TableAccent}{HTML}{E5F0FA}

\usepackage[utf8]{inputenc} 
\usepackage[T1]{fontenc}    
\usepackage{hyperref}       
\usepackage{url}            
\usepackage{booktabs}       
\usepackage{amsfonts}       
\usepackage{nicefrac}       
\usepackage{microtype}      
\usepackage{xcolor}         

\definecolor{linkcolor}{RGB}{255,0,0}
\definecolor{urlcolor}{RGB}{255,105,180}
\definecolor{citecolor}{RGB}{66,168,235}
\hypersetup{colorlinks=true,linkcolor=linkcolor,urlcolor=urlcolor,citecolor=blue}

\title{Omni-Customizer: End-to-End MultiModal Customization for Joint Audio-Video Generation}

\author{%
  Yuheng Chen$^{1}$\thanks{Equal contribution.} ~
  Qingdong He$^{3}$ \footnotemark[1] ~
  Teng Hu$^{1}$\footnotemark[1] ~
  Yuji Wang$^{1}$ ~
  Yabiao Wang$^{2}$ \\
  \textbf{Lizhuang Ma}$^{1}$\thanks{Corresponding author.} ~ 
  \textbf{Jiangning Zhang}$^{2}$\thanks{Project lead.}   \\
  \normalsize $^1$Shanghai Jiao Tong University ~~ $^2$Zhejiang University ~~ \\ $^3$University of Electronic Science and Technology of China \\
  {\tt\small Project Page: \url{https://aliothchen.github.io/projects/Omni-Customizer/}}
}

\begin{document}
\maketitle

\begin{figure}[htp]
    \centering
    \includegraphics[width=\linewidth]{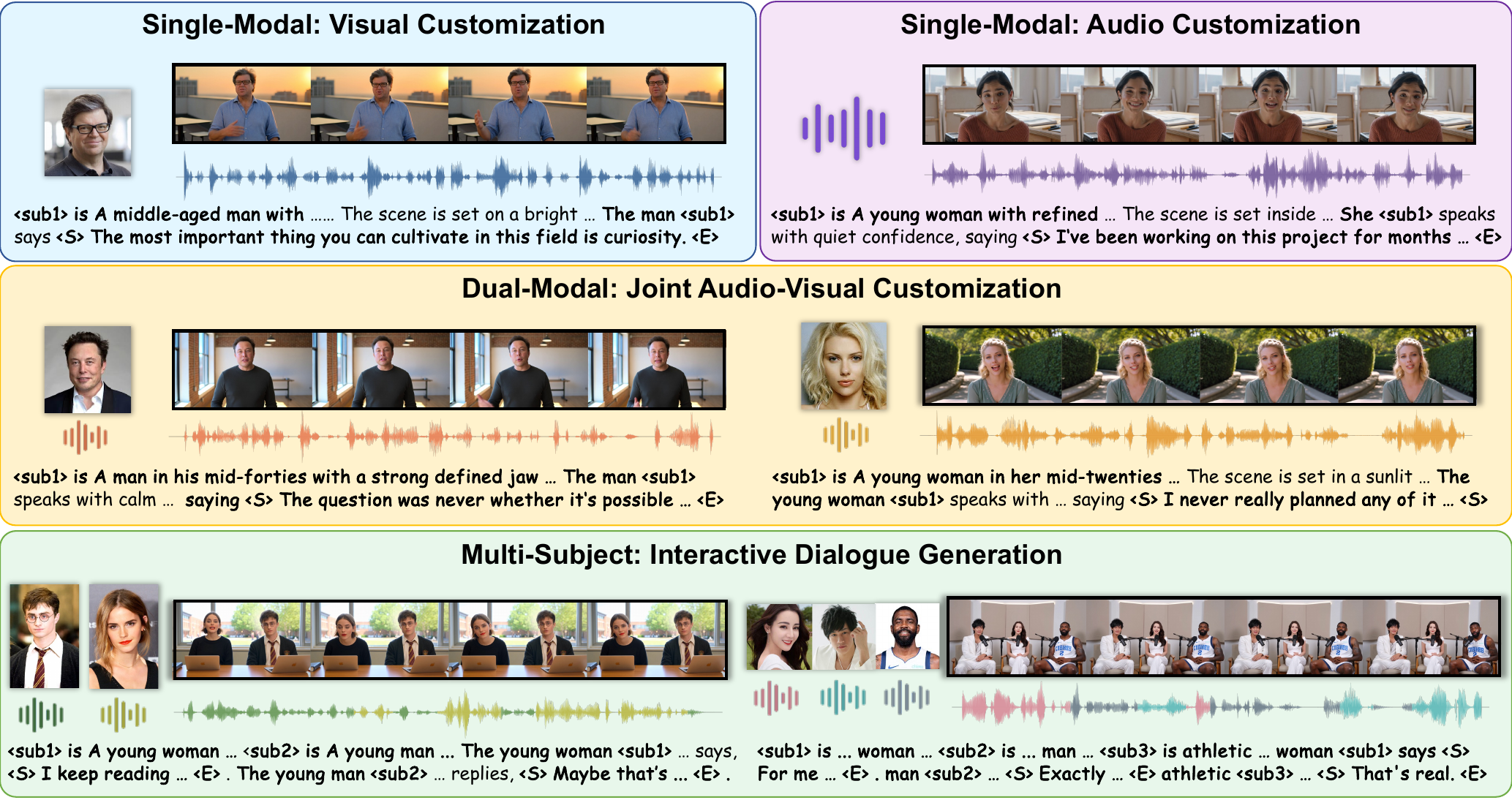}
    \caption{
Omni-Customizer achieves high-quality joint audio-video customization conditioned on \textbf{\textit{1)}} reference images, \textbf{\textit{2)}} reference audio, or \textbf{\textit{3)}} both. Furthermore, it demonstrates robust multimodal binding capabilities in  \textbf{\textit{4)}} highly realistic, multi-subject conversational scenarios.}
    \label{fig:teaser}
\end{figure}

\begin{abstract}
The landscape of joint audio and video generation has been fundamentally transformed by the advent of powerful foundation models. Despite these strides, achieving cohesive multimodal customization for the simultaneous preservation of visual identities and vocal timbres across multiple interacting subjects remains largely underexplored. 
To bridge this gap, we present Omni-Customizer, an end-to-end framework targeted at the precise binding and seamless fusion of multimodal identity information. 
Specifically, we introduce an Omni-Context Fusion (OCF) module that effectively enriches the base textual prompt with dense, multimodal identity cues, along with a Masked TTS Cross-Attention (MTP-CA) mechanism explicitly designed to prevent the severe "speech leakage" problem. 
Within this architecture, we propose Semantic-Anchored Multimodal RoPE (SA-MRoPE) to anchor visual and audio reference tokens, along with TTS embeddings, to their corresponding semantic descriptions, enabling structured multimodal fusion and robust identity binding.
Furthermore, we devise a comprehensive training strategy that incorporates interleaved audio-video scheduling to rapidly adapt the audio branch to multilingual scenarios without degrading foundational priors, and a progressive in-pair to cross-pair curriculum to facilitate the learning of high-level and robust identity features.
Extensive experiments demonstrate that Omni-Customizer achieves state-of-the-art performance in dual-modal customized generation, excelling across visual identity similarity, timbre consistency, precise audio-video synchronization, and overall video-audio fidelity.
\end{abstract}

\section{Introduction}
Following the open-source release of foundational models like Ovi~\cite{ovi} and LTX-2~\cite{ltx}, joint audio and video generation has garnered widespread attention within the research community. Recently, the remarkable success of proprietary models such as Seedance 2.0~\cite{seedance} has propelled joint synthesis to become nearly the default generative paradigm.
Despite these rapid advancements, open-source multimodal customization within this joint generation domain remains largely under-explored, especially in human-centric applications and complex interactive scenarios. 

To contextualize these challenges, existing customization efforts can generally be categorized into three paradigms. \textbf{\textit{1)}} First, while unimodal video customization~\cite{phantom,bindweave,kaleido,vace} and cross-modal driving pipelines~\cite{humo,wan-s2v,skyreels-a2} are highly mature, extending these systems to joint audio-visual generation requires non-trivial architectural changes (e.g., adding a separate audio tower and cross-modal coupling) that lie outside their original scope. \textbf{\textit{2)}} Second, although pioneering unified models like DreamID-Omni~\cite{dreamid-omni} support joint customization, the Syn-RoPE mechanism they devised fails to achieve robust cross-modal identity binding, making identity cues highly vulnerable to the rapid periodic decay of arbitrary positional offsets. \textbf{\textit{3)}} Finally, current joint frameworks built upon popular open-source backbones (e.g., Ovi~\cite{ovi}) suffer from inherent bottlenecks due to the limited speech reconstruction capacity of audio VAEs~\cite{mmaudio} and the unbalanced multilingual phonetic granularity of standard text encoders~\cite{umt5}. Moreover, Ovi is highly prone to a \emph{Caption Vocalization} anomaly, where the audio tower erroneously synthesizes non-speech descriptive captions into spoken audio. These inherent limitations hinder their deployment in complex, real-world interactive scenarios.

To overcome these fundamental limitations, we propose \textbf{\textit{Omni-Customizer}}, an end-to-end framework tailored for human-centric joint audio-video customization. To achieve efficient and precise multimodal identity binding, we first introduce the \textbf{Omni-Context Fusion (OCF)} module, which enriches the text representation with dense multimodal cues. For semantic-aware cross-modal fusion, we propose \textbf{Semantic-Anchored Multimodal RoPE (SA-MRoPE)}, featuring a unified 3D positional space that elegantly anchors disparate multimodal reference tokens directly to their corresponding subject descriptions. Additionally, to avert potential \emph{Caption Vocalization} anomalies, we employ a \textbf{Masked TTS Cross-Attention (MTP-CA)} mechanism to strictly confine phoneme injection within designated speech spans. Finally, to fully exploit available training datasets despite their severely skewed language distributions (e.g., predominantly Chinese data), we devise an \textbf{Interleaved Modality-Decoupled Training strategy}. By alternating between joint audio-video optimization and large-batch audio-only steps, this approach empowers the audio branch to rapidly acquire foundational multilingual capabilities without compromising the backbone's inherent lip-sync and cross-modal alignment priors. This is further complemented by a \textbf{progressive in-pair to cross-pair curriculum}, enabling the model to cultivate highly robust and high-level identity representations.

Extensive experiments on our newly proposed \textbf{O}mni-\textbf{C}ustomizer \textbf{Bench}mark (\textbf{OC-Bench}) validate the superiority of our framework. Omni-Customizer achieves exceptional single-modal video and audio quality, alongside fine-grained audio-video synchronization. Furthermore, it ensures robust dual-modal identity customization, enabling precise cross-modal binding and correspondence even in complex multi-subject scenarios, thereby cconfirming the efficacy of our innovations in architecture and training strategy.
In summary, our main contributions are as follows: \\
\hspace*{1em}\textit{\textbf{1)}} We propose Omni-Customizer, an end-to-end framework tailored for human-centric joint audio-visual customized generation. Specifically, we introduce the Omni-Context Fusion (OCF) module, which seamlessly enriches text representations with dense multimodal cues to achieve efficient and precise identity binding. \\
\hspace*{1em}\textit{\textbf{2)}} We design Semantic-Anchored Multimodal RoPE (SA-MRoPE), utilizing a unified 3D positional space to anchor reference tokens to their semantic descriptions, thereby resolving multi-subject identity confusion. Additionally, we incorporate a Masked TTS Cross-Attention (MTP-CA) mechanism to strictly confine phoneme injection and completely avert \emph{Caption Vocalization} anomalies. \\
\hspace*{1em}\textit{\textbf{3)}} We devise an Interleaved Modality-Decoupled Training strategy that empowers the model to rapidly acquire multilingual capabilities without compromising inherent alignment priors
. 
Paired with a progressive in-pair to cross-pair curriculum, this approach effectively cultivates robust, high-level identity representations. \\
\hspace*{1em}\textit{\textbf{4)}} We develop a comprehensive data curation pipeline, yielding a highly diverse multi-subject multimodal dataset and the comprehensive OC-Bench. Extensive evaluations demonstrate that Omni-Customizer achieves state-of-the-art performance across video and audio quality, precise audio-video synchronization, and dual-modal identity preservation.

\section{Related Works}
\subsection{Joint Audio-Video Generation}
The architectural transition from U-Net~\cite{unet,d-unet} to Diffusion Transformers (DiT)~\cite{dit} has catalyzed the emergence of powerful foundation models in both video~\cite{wan,hunyuanvideo,opensora} and audio~\cite{f5tts,hy-foley} generation. Leveraging these robust unimodal priors, subsequent works have advanced cross-modal generation, facilitating high-fidelity Audio-driven Video (A2V)~\cite{hunyuanvideo,humo,skyreels-a2,wan-s2v} and Video-to-Audio (V2A)~\cite{mmaudio,foleycrafter,diff-foley} synthesis. Recently, the field has reached a new milestone with the advent of native Joint Audio-Video Generation (JAVG). Advanced dual-stream DiT-based models~\cite{ovi,harmony,javisdit,universe,ltx} have established robust baselines for concurrent synthesis and garnered widespread attention. However, these frameworks predominantly focus on general-purpose content creation, remaining largely underexplored in complex scenarios requiring fine-grained control, such as multi-subject interactions and identity-preserving customization, thereby highlighting a critical gap in current generative capabilities.

\subsection{Video and Audio Customization}
Early U-Net-based explorations primarily adopted a decoupled paradigm for motion and appearance customization~\cite{videomage,decouplemotion,motiondirector}. As DiT took the lead, the field rapidly transitioned toward efficient end-to-end video customization frameworks for general subjects~\cite{phantom,kaleido,bindweave,firstframe,mulit-concept,movieweaver,dreambooth,videodreamer,conceptmaster}. Given human sensitivity to facial inconsistencies, a specialized line of work has focused exclusively on human-centric identity preservation~\cite{id-animator,consistid}, which specifically addresses the stringent requirements of maintaining high-fidelity identities across complex and dynamic scenarios. 
In parallel, audio customization has progressed rapidly through voice cloning and zero-shot multi-speaker TTS, enabling faithful speaker adaptation from short reference speech and extending to multilingual settings~\cite{qwen3-tts,transfervoice,valle}.
Despite these unimodal successes, concurrent identity customization across both audio and video remains highly underexplored, especially in multi-subject contexts. While recent bimodal explorations like DreamID-Omni~\cite{dreamid-omni} attempt to synchronize visual and vocal identities, their lack of deep multimodal binding poses significant challenges when confronted with multi-subject interactions. Addressing this unified alignment remains a critical gap that our work seeks to resolve.

\section{Data Curation}
\label{sec:data}

\noindent\textbf{Source Data Collection.}
We construct our customization-centric multi-subject audio-video dataset using OpenHumanVid~\cite{openhumanvid} and OpenS2V-5M~\cite{opens2v} as source corpora. We first remove clips lacking audio and filter the remaining videos based on metadata quality scores provided by respective datasets.\\
\noindent\textbf{Reference Image Extraction.}
Our extraction strategy is tailored to the source dataset type: \textbf{\textit{1)}} For OpenHumanVid (used primarily for in-pair generation), we run InsightFace~\cite{arcface,retinaface} face tracking on every clip, selecting the frame that maximizes the product of detection confidence and bounding box area as the reference image. \textbf{\textit{2)}} For the filtered subset of OpenS2V-5M (used for cross-pair generation), we leverage their provided subject reference images and spatially re-match them to their native InsightFace tracks via mask-level IoU~\cite{iou}.\\
\noindent\textbf{ASR and Audio Captioning.}
For each clip, we run Qwen3-Omni-30B-A3B~\cite{qwen3-omni} to produce timestamped ASR transcripts. Each segment is annotated with structural fields: \{\emph{speaker, text, start, end, language}\}. Simultaneously, the model generates a global audio caption that comprehensively summarizes the prosody and surrounding acoustic environment.\\
\noindent\textbf{Reference Audio Synthesis.}
To circumvent the in-pair copy-paste shortcut (as discussed in Sec.~\ref{sec:training}) and explicitly disentangle phonetic content from timbre, for each identified speaker, we extract their longest continuous audio segment and its corresponding ASR text to condition CosyVoice3~\cite{cosyvoice3} for re-synthesizing a reference audio clip, thereby producing a vocal exemplar that strictly matches the speaker's identity but neutralizes the surface acoustic and linguistic context.\\
\noindent\textbf{MLLM-guided Omni-Binding.}
The final critical step serves a dual purpose: \textbf{\textit{1)}} to structurally link the visual identity (FaceID) with the vocal identity (SpeakerID); \textbf{\textit{2)}} to generate the semantically anchored structured captions required by our OCF module and the Ovi backbone. To achieve this, both MLLMs are provided with the source audio-video clip, ASR transcripts, and candidate reference image and audio pools to simultaneously output the exact identity binding and the semantically anchored prompt. Specifically, we adopt a routed ensemble strategy: \textbf{\textit{1)}} \textbf{Gemini~2.5-Pro}~\cite{gemini2.5} handles potential multi-person interacting scenarios ($\#\text{faces}{>}1$ or $\#\text{speakers}{>}1$), and \textbf{\textit{2)}} \textbf{Qwen3-Omni-30B-A3B} processes the bulk of straightforward scenes ($\#\text{faces}{\le}1$ and $\#\text{speakers}{\le}1$).

\begin{figure}[t]
    \centering
    \includegraphics[width=\linewidth]{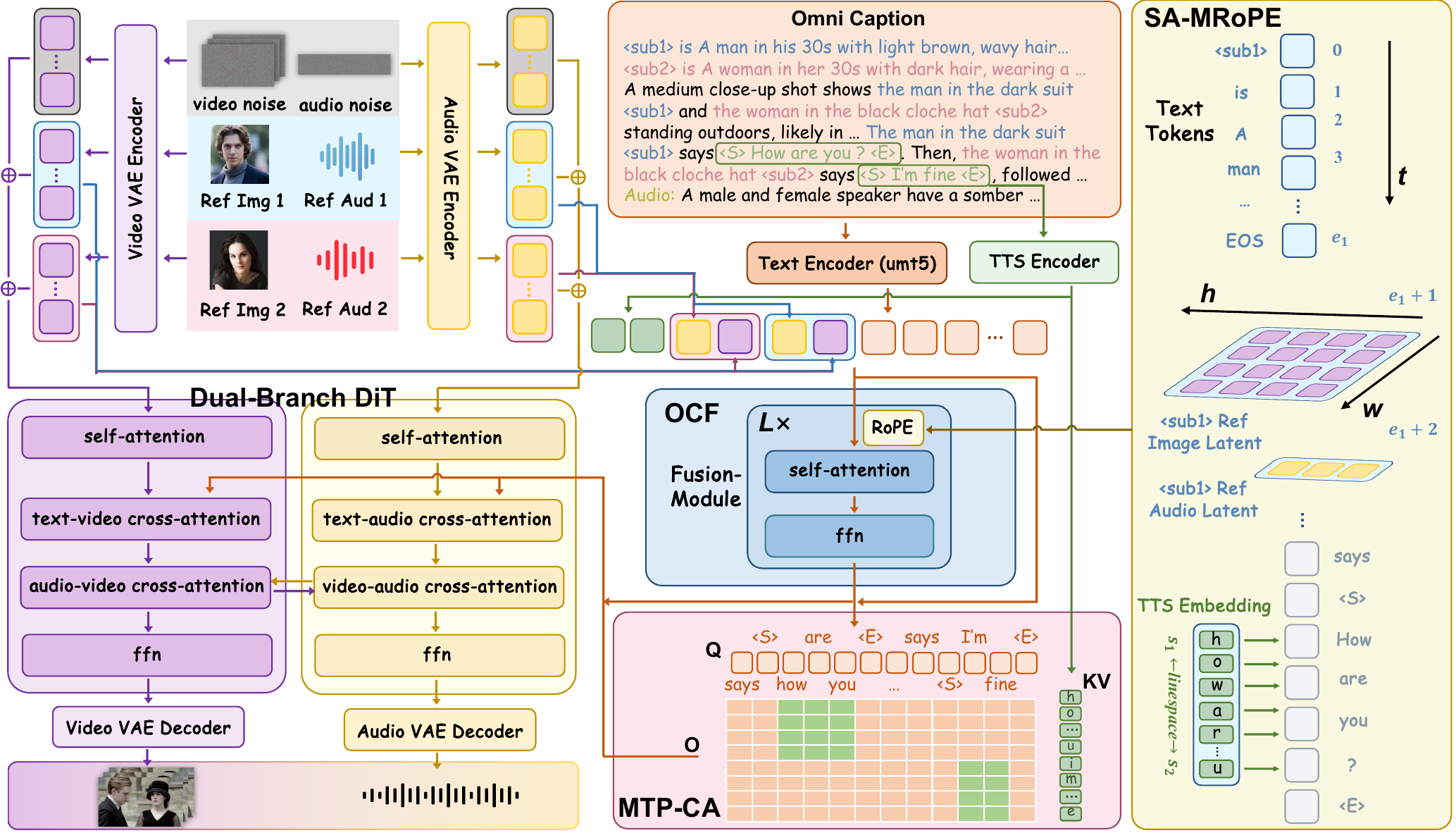}
    \caption{Framework of Omni-Customizer: The text prompt, TTS embeddings, reference images, and audios are integrated by the OCF module, which employs SA-MRoPE for precise multimodal binding. Additionally, MTP-CA ensures exclusive pronunciation enhancement for spoken texts. The enhanced context is injected into the dual-stream backbone for cohesive multi-subject identity preservation.}
    \label{fig:method}
    \vspace{-10pt}
\end{figure}

\section{Method}
\label{method}
\subsection{Formulation of Joint Audio-Video Customization}
\label{subsec:framework_overview}

\noindent\textbf{Symmetric Dual-Stream Architecture.} Omni-Customizer is built upon a dual-stream Diffusion Transformer (DiT) architecture, initialized directly from the pre-trained Ovi~\cite{ovi} backbone. Formally, given a sequence of subject reference images $\mathcal{I} = \{I_1, \dots, I_N\}$, corresponding reference audios $\mathcal{A} = \{A_1, \dots, A_N\}$, and a text prompt $P$, our framework aims to jointly generate the customized video and audio target latents for the final audio-video output.\\
During the diffusion denoising process at timestep $t$, let $z_{v,t}$ and $z_{a,t}$ denote the noisy target latents for the video and audio modalities. To explicitly condition the generation, the video and audio references are pre-encoded into latent representations $c_v$ and $c_a$, respectively, while the text prompt yields the text embedding $c_{txt}$. Therefore, the joint denoising process of our dual-stream DiT, denoted as $\mathcal{F}_\theta$, is elegantly formulated as a unified forward pass:
\begin{equation}
    (\hat{\epsilon}_{v}, \hat{\epsilon}_{a}) = \mathcal{F}_\theta \Big( [z_{v,t} \oplus c_v], [z_{a,t} \oplus c_a], t, c_{txt} \Big)
\end{equation}
where $\oplus$ denotes concatenation along the spatial and temporal sequence dimensions, and $(\hat{\epsilon}_{v}, \hat{\epsilon}_{a})$ represents the joint model predictions (e.g., velocity or noise) for both modalities. Guided by the multimodal text embedding $c_{txt}$, this unified formulation seamlessly translates the aligned reference priors into the target generation space.

\noindent\textbf{Structured Omni-Caption.} To leverage the strong text-following capabilities of the Ovi backbone, we utilize MLLMs to re-caption the data into a standardized format (detailed in Sec.~\ref{sec:data}). Formally, the constructed prompt $P$ is defined as:
\begin{equation}
\begin{aligned}
    & \underbrace{L_1\texttt{ <sub1> is } D_{v,1}\texttt{, with } D_{a,1}\texttt{.}}_{\substack{P_{sub,1} \\ \text{\scriptsize Subject 1 Descriptor}}} \dots 
      \underbrace{L_N\texttt{ <subN> is } D_{v,N}\texttt{, with } D_{a,N}\texttt{.}}_{\substack{P_{sub,N} \\ \text{\scriptsize Subject N Descriptor}}} \\
    & \underbrace{D_{env}\texttt{ } D_{act}(\dots L_i\texttt{ <sub}_i\texttt{> acts} \dots)}_{\substack{P_{vid} \\ \text{\scriptsize Global Environment and Action}}} \ \ 
      \underbrace{L_k\texttt{ <sub}_k\texttt{> says <S> } T_{k,j} \texttt{ <E>.}}_{\substack{P_{speech} \\ \text{\scriptsize Speech Content}}}
\end{aligned}
\end{equation}
where $P_{sub,i}$ denotes the multimodal descriptor for the $i$-th subject. Within this descriptor, $L_i$ represents a natural, distinctive identity label (e.g., ``the man in red'') prepended to the anchor token \texttt{<sub}$_i$\texttt{>}. This design preserves the semantic integrity of the prompt, making it easier for the text encoder to comprehend without disrupting its pre-trained natural language distribution. $D_{v,i}$ and $D_{a,i}$ represent the explicit visual and acoustic descriptions for the $i$-th subject. The terms $D_{env}$ and $D_{act}$ jointly constitute a standard Text-to-Video (T2V) prompt, depicting the global environment and overall actions, but with the subjects persistently referenced via their anchor tokens. $T_{k,j}$ denotes the $j$-th spoken utterance of the active speaker $k$, strictly enclosed by the speech markers \texttt{<S>} and \texttt{<E>}. 
By design, the anchor token \texttt{<sub}$_i$\texttt{>} seamlessly connects the diverse cross-modal semantics (i.e., visual appearance, acoustic timbre, physical action, and spoken text) belonging to the exact same subject throughout the entire prompt.

\subsection{Omni-Context Fusion and Semantic Anchoring}
\label{subsec:ocf_and_anchoring}

Simply depending on textual features to bind the multimodal identity conditions ($c_v$ and $c_a$) to the appropriate spatiotemporal regions of the target latents ($z_{v,t}$ and $z_{a,t}$) is highly unreliable. In vanilla DiT architectures, the text embeddings, video reference latents, and audio reference latents never interact simultaneously within a unified module. Instead, they only interact indirectly through the noisy target latents during denoising, typically by independently injecting modality-specific hints into the main denoising stream. To achieve precise cross-modal alignment and deep identity binding, we design a comprehensive multimodal prompt enrichment and conditioning pipeline.

\noindent\textbf{Omni-Context Fusion (OCF).} 
Rather than relying on the diffusion backbone to resolve complex multimodal alignments, we propose OCF to elevate the foundational text encoder~\cite{umt5} into an active cross-modal alignment engine. Specifically, we concatenate the base text embeddings $c_{txt}$, the visual reference tokens $c_v$, the audio reference tokens $c_a$, and the supplementary TTS phoneme embeddings $c_{tts}$ into a unified input sequence, denoted as $S = [c_{txt} \oplus c_v \oplus c_a \oplus c_{tts}]$. 
The inclusion of $c_{tts}$, which is encoded via F5-TTS~\cite{f5tts} from the spoken text enclosed by \texttt{<S>} and \texttt{<E>}, acts as a crucial phonetic bridge, explicitly aligning the textual spoken content with the acoustic timbre prior. 
This combined sequence is then iteratively processed through $L$ dedicated transformer blocks to enforce deep cross-modal interaction. To absorb the multimodal context while preserving the integrity of the pre-trained language representations, at each layer, we extract the first $\text{len}(c_{txt})$ tokens of the output and add them back to the original $c_{txt}$ as a residual connection~\cite{residual}. We apply zero-initialization to the projection layers of these residuals to ensure they are strictly zero at the start of training, guaranteeing overall optimization stability. Through the OCF module, the text embeddings are enriched with dense cross-modal awareness, which significantly facilitates the precise binding and injection of identity information in the subsequent DiT blocks.

\noindent\textbf{Semantic Anchored Multimodal RoPE (SA-MRoPE).} 
While the OCF module aggregates multimodal inputs into a unified sequence, treating these heterogeneous tokens uniformly without structural distinction is highly suboptimal. 
Specifically, text tokens are naturally organized as one-dimensional sequences, whereas image tokens exhibit a two-dimensional spatial structure, and audio features possess their own temporal dynamics. This inherent structural mismatch hinders the precise alignment and fusion of information across modalities, leading to ineffective interaction modeling and potential identity entanglement.
To facilitate more effective cross-modal interaction while preserving the intrinsic semantics of each modality, we introduce SA-MRoPE which explicitly anchors the multimodal reference tokens to their corresponding semantic subject descriptions within the text sequence in a structured and modality-position-aware manner. 
Formally, for a given subject $k$ in the prompt, let its corresponding descriptor $P_{sub,k}$ span the 1D temporal token indices $[s_k, e_k]$. We assign the 3D positional coordinates for its associated visual reference tokens $Z_{img}^{(k)}$ and audio reference tokens $Z_{aud}^{(k)}$ as follows:
\begin{equation}
    Pos(Z_{img}^{(k)}) = (e_k + 1, h, w), \quad Pos(Z_{aud}^{(k)}) = (e_k + 2, j, 0)
\end{equation}
where $h$ and $w$ are the spatial coordinates of the visual reference tokens, and $j$ is the temporal sequence index of the audio reference tokens. Subsequent text tokens in the prompt resume their temporal positions starting from $e_k + 3$. \\
For the TTS phoneme tokens $Z_{tts}^{(k)}$, we map their positions directly onto the semantic speech content span $[t_{start}, t_{end}]$ determined by the \texttt{<S>} and \texttt{<E>} tags using linear interpolation. Crucially, we set the final coordinate dimension to $1$ to explicitly distinguish these synthetic phoneme tokens from the base prompt text embeddings (which default to $0$ in this dimension):
\begin{equation}
    Pos(Z_{tts}^{(k)}) = (\text{linspace}(t_{start}, t_{end}, \text{len}(Z_{tts}^{(k)})), 0, 1)
\end{equation}
This semantic anchoring naturally creates a strong spatial-temporal attention bias during the OCF forward pass, ensuring that each reference modality is rigidly bound to its correct textual identity without relying on arbitrary fixed offsets.

\noindent\textbf{Masked TTS-to-Prompt Cross-Attention (MTP-CA).} 
While the OCF module enriches the prompt and SA-MRoPE provides an effective spatial-temporal attention bias, they guide the cross-modal interaction in a soft manner rather than imposing strict isolation constraints. Consequently, the framework remains susceptible to an anomaly inherent to the pre-trained Ovi backbone, where non-speech descriptive content inadvertently leaks into the generated audio stream, a phenomenon we term \textit{Caption Vocalization} (further detailed in the  supplementary material). 
Since the audio tower processes the entire text prompt globally, it relies heavily on the \texttt{<S>} and \texttt{<E>} tokens to demarcate speech. While these embeddings provide a baseline boundary signal, such token-level soft constraints can occasionally be overwhelmed in complex, information-dense multi-subject prompts. 
To surgically resolve this anomaly, we propose MTP-CA, which bridges the prompt embeddings $c_{txt}$ and the TTS phoneme embeddings $c_{tts}$ via a masked cross-attention mechanism. Specifically, we inject these phoneme priors strictly into the text tokens located within the \texttt{<S>...<E>} span. A binary mask ensures that all non-speech narrative regions receive exactly zero phoneme-level excitation. Consequently, the audio tower receives precise pronunciation and acoustic guidance exclusively for the intended dialogue. This hard-gating strategy completely eradicates Caption Vocalization while simultaneously endowing the framework with robust multilingual speech capabilities.

\subsection{Training Strategy}
\label{sec:training}


\noindent\textbf{Interleaved JAVG and TTS-only Steps.}
The pre-trained Ovi backbone relies predominantly on English corpora~\cite{ovi}, leaving a large portion of the OpenHumanVid and OpenS2V datasets highly out-of-distribution (OOD). Simply fine-tuning on this data risks inadvertently degrading the model's native lip-sync capabilities. This risk is further amplified by the \textit{suboptimal reconstruction capability} of the MMAudio VAE~\cite{mmaudio}, particularly for human speech. Additionally, since the number of audio tokens is significantly smaller than that of video tokens, direct joint training inevitably leads to \textit{an unbalanced optimization of the audio branch} (refer to the supplementary material). 
To fully utilize the training datasets and rapidly adapt the model to the complex OOD speech domain without sacrificing its original multimodal alignment, we alternate two step types during training: \\
\hspace*{1em} \textit{\textbf{1)}} \textbf{JAVG step} (ratio $r$): Joint forward pass of both DiTs with multimodal cross-attention enabled to optimize complete cross-modal feature alignment. \\
\hspace*{1em} \textit{\textbf{2)}} \textbf{TTS-only step} (ratio $1{-}r$): Forward pass of only the audio DiT. The multimodal cross-attention target is null, rendering the cross-modal gradient pathway structurally inactive. \\
This interleaved strategy benefits training in two pivotal ways. \textbf{\textit{1)}} First, by substantially expanding the audio batch size during the TTS-only steps, we effectively average the influence of the MMAudio VAE reconstruction error on the training loss toward zero, ensuring an unbiased gradient estimate ideal for stable optimization. \textbf{\textit{2)}} Second, from a parameter update perspective, the TTS-only step plays a regularization role analogous to LoRA~\cite{lora}, since it freezes the cross-modal pathway, expanding the intra-modal audio capacity to assimilate new multilingual contexts and complex conversational dynamics, while protecting the already-learned audio-video interface. The interleaved JAVG steps then act as rehearsal, pulling the audio representations back to the expected input distribution and preventing the internal covariate drift that pure audio-only training would otherwise induce (see the supplementary material for detailed mathematical derivations).

\noindent\textbf{Progressive Disentanglement Curriculum.}
To thoroughly disentangle specific spoken content from acoustic timbre, we take a data-driven approach by synthesizing reference audio with randomized text via CosyVoice-3~\cite{cosyvoice3}. Concurrently, to mitigate the trivial copy-paste~\cite{phantom,phantom-data} shortcut and compel the model to learn high-level, robust visual representations, we curate a diverse reference image pool for each identity based on OpenS2V. However, we found that directly initiating training with complex multi-subject interactions under these strict disentanglement constraints potentially leads to catastrophic convergence failure. To achieve both ends stably, we propose a progressive two-stage curriculum: \\
\hspace*{1em} \textbf{\textit{1)}} \textbf{Stage A: Single-Subject Alignment.} We predominantly utilize in-pair data from OpenHumanVid, restricting the training to single-identity scenes. This simplified setting allows the model to rapidly adapt to the newly introduced architecture and acquire basic customization capabilities.\\
\hspace*{1em} \textbf{\textit{2)}} \textbf{Stage B: Multi-Subject Disentanglement.} We escalate to complex multi-subject training using the cross-pair data from OpenS2V. By completely decoupling the references from the target generation, this stage endows the model with advanced multi-subject customization skills and forces the extraction of intrinsic, abstract multimodal identity features.

\begin{figure}[t]
    \centering
    \includegraphics[width=\linewidth]{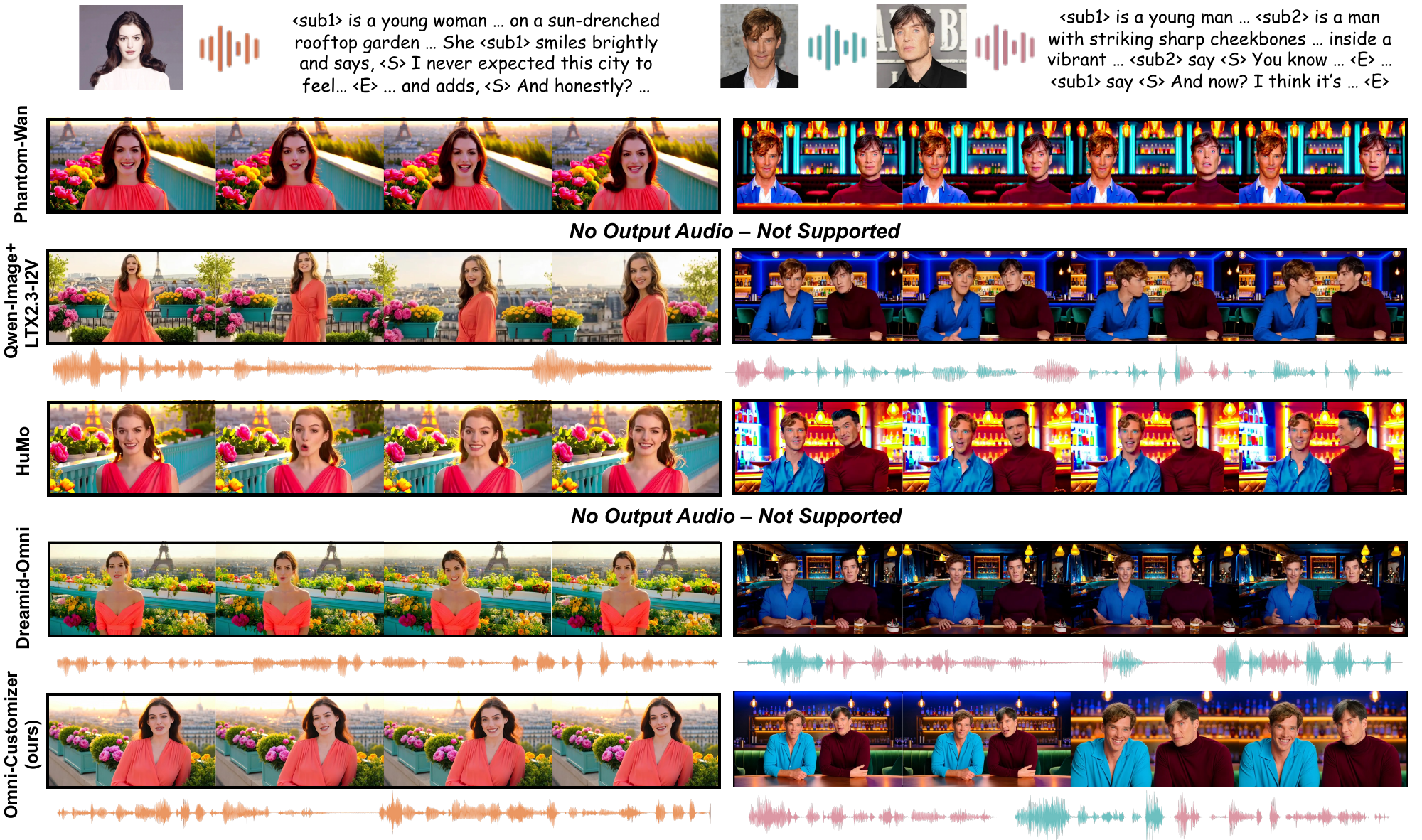}
    \caption{Qualitative comparison with state-of-the-art baselines chosen from four different paradigms.}
    \label{fig:compare}
    \vspace{-10pt}
\end{figure}

\section{Experiments}
\label{sec:exp}
\subsection{Experimental Details}
\label{sec:implementation}
\noindent\textbf{Training details.}
We initialize our Omni-Customizer directly from the pre-trained Ovi backbone~\cite{ovi}. Following the training strategies outlined in Sec.~\ref{sec:training}, our progressive training process is structured into three distinct stages:
\noindent\textbf{\textit{1) Stage 1: Single-Subject Alignment and Audio Bootstrapping (20K steps).}} The model is audio-video joint trained on 0.7M single-subject aesthetically filtered in-pair clips from the OpenHumanVid dataset with a batch size of 64, interleaved with TTS-only steps trained on the Emilia dataset~\cite{emilia} with a batch size of 1024. The step ratio between JAVG and TTS-only optimization is set to 1:1. 
\noindent\textbf{\textit{2) Stage 2: Multi-Subject Adaptation (10K steps).}} The model is adapted to multi-subject scenarios on the 0.3M multi-subject OpenHumanVid subset for 10K steps using exclusively the JAVG steps with a batch size of 64.
\noindent\textbf{\textit{3) Stage 3: Cross-Pair Disentanglement (10K steps).}} To achieve robust and high-level identity disentanglement, the model continues to audio-video joint train on a 0.5M subset from the OpenS2V dataset with a batch size of 64. 
For optimization, all stages are optimized using AdamW ($\beta_1=0.9, \beta_2=0.95$) with a weight decay of $0.01$. We employ a cosine learning rate scheduler specifically applied to the newly added OCF and MTP-CA, which gradually decays from an initial learning rate of 1e-4 down to 1e-5 finally. 

\noindent\textbf{OC-bench and metrics.}
\label{sec:metrics}
To facilitate a rigorous evaluation of multimodal customization, we introduce the \textbf{O}mni-\textbf{C}ustomizer \textbf{Bench}mark (\textbf{OC-Bench}), a comprehensive benchmark consisting of 300 test cases structured into three 100-item subsets of escalating complexity:
\noindent\textbf{\textit{1) Single-Subject Customization.}} Evaluates basic joint audio-visual customization capabilities using single-speaker prompts.
\noindent\textbf{\textit{2) Robust Identity Binding.}} Assesses multimodal binding robustness within standard two-person dialogue scenarios.
\noindent\textbf{\textit{3) Multi-Subject Complex Scenes.}} Features more challenging cases involving off-screen speakers, silent identities, and multilingual dialogue. 
\noindent We employ a streamlined suite of automated metrics: 
\noindent\textbf{\textit{1) Identity Preservation:}} Face Similarity and temporal Face Consistency (ArcFace~\cite{arcface}); Timbre Similarity (T-Sim via WavLM~\cite{wavlm}).
\noindent\textbf{\textit{2) AV-Sync:}} Lip-sync accuracy (Sync-C, Sync-D)~\cite{syncnet}; IB-Score~\cite{imagebind}.
\noindent\textbf{\textit{3) Video Quality:}} Aesthetic Quality (Aesthetic-v2.5)~\cite{laion}; Imaging Quality (MUSIQ)~\cite{musiq}; Temporal Flickering~\cite{vbench}.
\noindent\textbf{\textit{4) Audio Quality:}} AudioBox-Aesthetics (PQ)~\cite{audiobox}; Word Error Rate (WER, Whisper-v3)~\cite{wer}; IB-A Score~\cite{imagebind}.

\definecolor{TableAccent}{HTML}{E5F0FA}
\begin{table*}[t]
\centering
\caption{Quantitative comparison with state-of-the-art methods on OC-Bench. \textbf{Bold} and \underline{underline} represent the best and second-best results, respectively.}
\label{tab:main_results}
\resizebox{\linewidth}{!}{
\setlength{\tabcolsep}{4.2pt} 
\begin{tabular}{l cc cc ccc ccc}
\toprule
\multirow{2}{*}{\textbf{Method}} & \multicolumn{2}{c}{\textbf{Identity Preservation}} & \multicolumn{2}{c}{\textbf{AV-Sync}} & \multicolumn{3}{c}{\textbf{Video Quality}} & \multicolumn{3}{c}{\textbf{Audio Quality}} \\
\cmidrule(lr){2-3} \cmidrule(lr){4-5} \cmidrule(lr){6-8} \cmidrule(lr){9-11}
& Face-Sim$\uparrow$/Cons$\uparrow$ & T-Sim$\uparrow$ & Sync-C$\uparrow$/D$\downarrow$ & IB-S$\uparrow$ & AQ$\uparrow$ & IQ$\uparrow$ & TF$\uparrow$ & WER$\downarrow$ & PQ$\uparrow$ & IB-A$\uparrow$ \\
\midrule
Phantom~\cite{phantom} & 0.657 / 0.882 & - & - / - & - & 0.322 & 0.431 & 0.853 & - & - & - \\
VACE~\cite{vace} & 0.674 / 0.895 & - & - / - & - & 0.345 & 0.534 & 0.862 & - & - & - \\
\midrule
Humo~\cite{humo} & 0.708 / 0.941 & - & 3.421 / 10.23 & 0.124 & 0.521 & 0.612 & 0.887 & - & - & - \\
HunyuanCustom~\cite{hunyuancustom} & 0.732 / 0.954 & - & 3.752 / 9.842 & 0.181 & 0.574 & \underline{0.654} & 0.908 & - & - & - \\
Wan2.2-S2V~\cite{wan-s2v} & 0.774 / 0.963 & - & 5.864 / 8.521 & 0.122 & 0.518 & 0.642 & \underline{0.954} & - & - & - \\
SkyReel-A2~\cite{skyreels-a2} & 0.761 / 0.958 & - & 4.218 / 9.124 & 0.184 & 0.552 & 0.638 & 0.941 & - & - & - \\
\midrule
Universe-1~\cite{universe} & 0.642 / 0.912 & - & 5.012 / 9.421 & 0.076 & 0.412 & 0.574 & 0.842 & 0.431 & 3.41 & 0.072 \\
Ovi~\cite{ovi} & 0.692 / 0.934 & - & 5.421 / 8.942 & 0.084 & 0.435 & 0.592 & 0.864 & 0.342 & 3.64 & 0.084 \\
MOVA~\cite{mova} & 0.695 / 0.936 & - & 5.425 / 8.938 & 0.085 & 0.438 & 0.594 & 0.866 & 0.338 & 3.65 & 0.086 \\
LTX2.3~\cite{ltx} & 0.742 / 0.952 & - & 6.028 / 8.214 & 0.092 & 0.484 & \textbf{0.672} & 0.878 & \underline{0.224} & 3.92 & 0.098 \\
\midrule
DreamID-Omni~\cite{dreamid-omni} & \underline{0.789} / \underline{0.967} & \underline{0.471} & \underline{6.082} / \underline{8.024} & \underline{0.188} & \underline{0.584} & 0.648 & 0.945 & 0.284 & \underline{4.12} & \underline{0.112} \\
\rowcolor{TableAccent} 
\textbf{Omni-Customizer (Ours)} & \textbf{0.812} / \textbf{0.976} & \textbf{0.514} & \textbf{6.235} / \textbf{7.821} & \textbf{0.194} & \textbf{0.592} & \underline{0.654} & \textbf{0.968} & \textbf{0.152} & \textbf{4.32} & \textbf{0.124} \\
\bottomrule
\end{tabular}
}

\end{table*}

\subsection{Comparisons and Analysis}
\label{sec:comparison}
To comprehensively evaluate Omni-Customizer, we compare it against leading state-of-the-art models on OC-Bench across four distinct paradigms:
\textit{\textbf{1)}} Video Customization, including Phantom~\cite{phantom} and VACE~\cite{vace}. We exclusively evaluate visual customization and identity preservation as these models lack native audio-generation capabilities.
\textit{\textbf{2)}} Audio-Driven Video Customization, including Humo~\cite{humo}, HunyuanCustom~\cite{hunyuancustom}, Wan2.2-S2V~\cite{wan-s2v} and SkyReel-A2~\cite{skyreels-a2}. We evaluate video quality and AV-sync but omit audio metrics, as the driving audio is a fixed input condition rather than a generative output.
\textit{\textbf{3)}} Qwen-Image + JAVG Models. We generate the first frame using Qwen-Image~\cite{qwen-image}, and then baselines (Ovi~\cite{ovi}, LTX2.3~\cite{ltx}, Universe~\cite{universe}, and MOVA~\cite{mova}) generate the video in an I2V manner.
\textit{\textbf{4)}} Joint Audio-Video Customization. Evaluates end-to-end unified multimodal customization, including DreamID-Omni~\cite{dreamid-omni}.

\noindent\textbf{Quantitative analysis.} As shown in Tab.~\ref{tab:main_results}, Omni-Customizer outperforms all baselines across core multimodal metrics. While video-only methods and cascaded pipelines (e.g., LTX2.3) maintain competitive general video quality (AQ/IQ), they suffer from poor identity binding and consistency. In contrast, our model achieves a significant lead in Face-Sim and T-Sim, demonstrating superior visual and acoustic fidelity. Notably, as complexity increases in Subsets 2 and 3, baselines experience sharp performance drops due to identity interference and sync failures. Our approach remains robust, maintaining high IB-Score and the lowest WER and Sync-D, effectively handling the challenges of multi-subject interaction and cross-modal alignment.

\noindent\textbf{Qualitative analysis.} As illustrated in Fig.~\ref{fig:compare}, we compare Omni-Customizer with state-of-the-art baselines. Phantom exhibits facial rigidity in two-subject scenarios. LTX2.3 suffers from gradual identity drift in subsequent frames. HuMo struggles with identity preservation in dual-person customization, showing mediocre consistency. DreamID-Omni performs suboptimally in both visual and acoustic modalities, resulting in noticeable identity entanglement and drift.
In contrast, Omni-Customizer achieves high-fidelity customization across both visual and acoustic modalities. Our model maintains robust identity binding and stable multi-subject consistency even in complex scenes, ensuring precise lip-sync without identity confusion.

\begin{figure}[t]
    \centering
    \includegraphics[width=\linewidth]{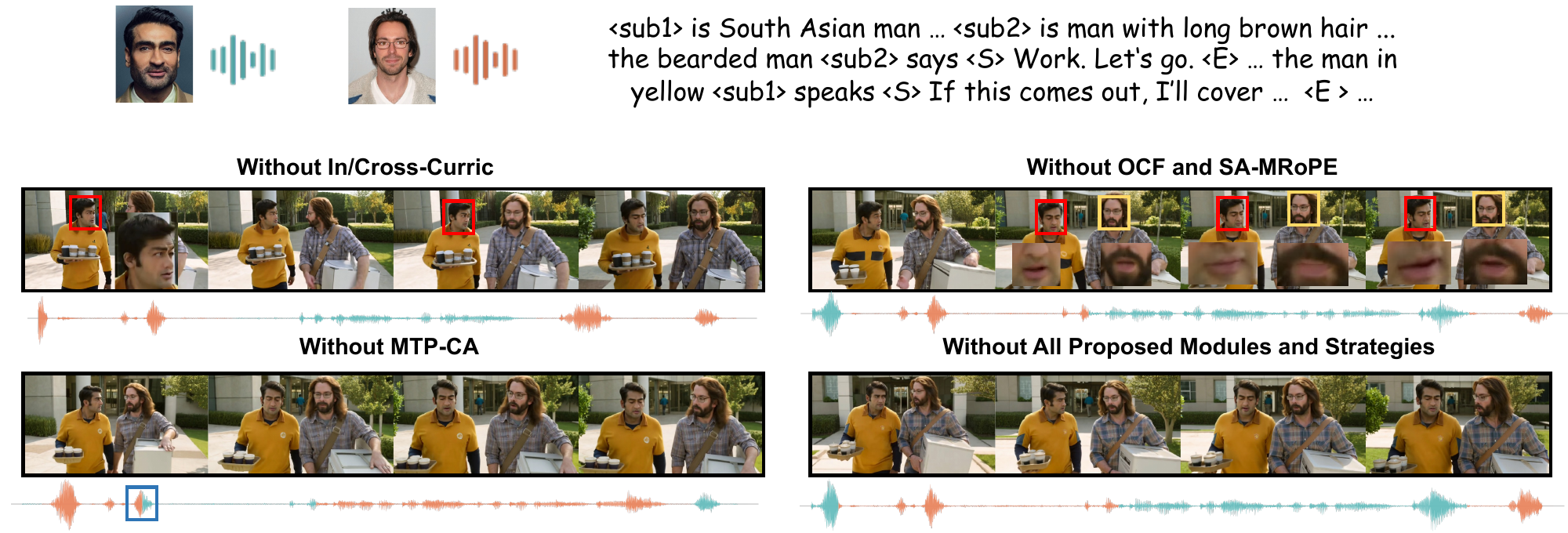}
    \caption{Qualitative ablation study of proposed modules and strategies.}
    \label{fig:ablation}

\end{figure}

\noindent\textbf{Ablation study.} Tab.~\ref{tab:ablation} validates the contribution of each proposed component on OC-Bench. While the OCF module establishes a cohesive multimodal latent space, the addition of SA-MRoPE explicitly anchors reference latents to semantic text tokens, significantly boosting identity preservation. Furthermore, the MTP-CA mechanism substantially improves audio-visual synchronization and speech fidelity. Finally, TTS-interleaved training enhances general audio quality, while progressive curriculum learning guarantees robust feature decoupling in complex multi-subject scenarios. These quantitative gains are strongly corroborated by the qualitative results in Fig.~\ref{fig:ablation}. Specifically, without the progressive curriculum learning, the generated faces often exhibit distorted and rigid artifacts. Removing OCF and SA-MRoPE disrupts spatial-temporal alignment, causing severe confusion where two subjects erroneously speak simultaneously. Lastly, without MTP-CA, non-speech narrative captions inadvertently leak into the generated spoken audio stream. Specifically, the audio tower fails to isolate the speech span, causing the subject to erroneously vocalize structural tags or physical descriptors rather than delivering the intended dialogue. This anomalous \textit{Caption Vocalization} severely disrupts the conversational immersion and phonetic purity. These compounding improvements confirm that structured alignment is strictly required; our carefully designed modules work in synergy to enforce absolute semantic boundaries and eradicate cross-modal feature bleeding.

\definecolor{TableAccent}{HTML}{E5F0FA}
\begin{table*}[t]
\centering
\caption{Quantitative ablation study on OC-Bench. We progressively integrate proposed modules to evaluate their individual contributions. \textbf{Bold} and \underline{underline} denote the best and second-best results, respectively.}
\label{tab:ablation}
\resizebox{\linewidth}{!}{
\setlength{\tabcolsep}{3.8pt} 
\begin{tabular}{@{} ccccc | cc cc ccc ccc @{}} 
\toprule
\multirow{2}{*}{\textbf{OCF}} & \multirow{2}{*}{\shortstack{\textbf{SA-}\\\textbf{MRoPE}}} & \multirow{2}{*}{\shortstack{\textbf{MTP-}\\\textbf{CA}}} & \multirow{2}{*}{\shortstack{\textbf{Inter-}\\\textbf{TTS}}} & \multirow{2}{*}{\shortstack{\textbf{In/Cross-}\\\textbf{Curric.}}} & \multicolumn{2}{c}{\textbf{Identity Preservation}} & \multicolumn{2}{c}{\textbf{AV-Sync}} & \multicolumn{3}{c}{\textbf{Video Quality}} & \multicolumn{3}{c}{\textbf{Audio Quality}} \\
\cmidrule(lr){6-7} \cmidrule(lr){8-9} \cmidrule(lr){10-12} \cmidrule(lr){13-15}
& & & & & Face-Sim/Cons$\uparrow$ & T-Sim$\uparrow$ & Sync-C/D$\uparrow \downarrow$ & IB-S$\uparrow$ & AQ$\uparrow$ & IQ$\uparrow$ & TF$\uparrow$ & WER$\downarrow$ & PQ$\uparrow$ & IB-A$\uparrow$ \\
\midrule
& & & & & 0.612 / 0.894 & 0.362 & 3.125 / 11.42 & 0.064 & 0.312 & 0.425 & 0.824 & 1.342 & 3.25 & 0.052 \\
\checkmark & & & & & 0.684 / 0.925 & 0.415 & 4.214 / 10.15 & 0.082 & 0.428 & 0.541 & 0.842 & 0.856 & 3.52 & 0.071 \\
\checkmark & \checkmark & & & & 0.742 / 0.948 & 0.458 & 4.862 / 9.241 & 0.135 & 0.512 & 0.594 & 0.882 & 0.642 & 3.82 & 0.094 \\
\checkmark & \checkmark & \checkmark & & & 0.765 / 0.958 & 0.482 & \underline{6.142} / \underline{7.954} & 0.162 & 0.554 & 0.612 & 0.914 & \underline{0.182} & 4.15 & 0.108 \\
\checkmark & \checkmark & \checkmark & \checkmark & & \underline{0.785} / \underline{0.965} & \underline{0.495} & 6.012 / 8.124 & \underline{0.181} & \underline{0.572} & \underline{0.638} & \underline{0.942} & 0.201 & \underline{4.24} & \underline{0.115} \\
\rowcolor{TableAccent}
\checkmark & \checkmark & \checkmark & \checkmark & \checkmark & \textbf{0.812} / \textbf{0.976} & \textbf{0.514} & \textbf{6.235} / \textbf{7.821} & \textbf{0.194} & \textbf{0.592} & \textbf{0.654} & \textbf{0.968} & \textbf{0.152} & \textbf{4.32} & \textbf{0.124} \\
\bottomrule
\end{tabular}
}

\end{table*}

\section{Conclusion}
\label{sec:conclusion}

In this paper, we propose Omni-Customizer, a novel end-to-end framework tackling cohesive multimodal customization in joint audio-video generation. To simultaneously preserve multi-subject visual identities and vocal timbres, we introduced Omni-Context Fusion (OCF) and Semantic-Anchored Multimodal RoPE (SA-MRoPE) for precise identity binding, alongside Masked TTS Cross-Attention (MTP-CA) to effectively mitigate speech leakage. Coupled with an interleaved, progressive training curriculum, Omni-Customizer achieves state-of-the-art performance in video fidelity, audio quality, and cross-modal consistency. 
Despite these successes, current generations are bounded to 720P resolution and 10-second durations. Scaling to higher resolutions and longer sequences presents profound challenges for both model architecture and the data curation pipeline, particularly in maintaining long-term identity consistency. Addressing these temporal and spatial scaling bottlenecks remains our primary focus for future work.







\bibliographystyle{plainnat}
\bibliography{references}


\appendix



\clearpage

\end{document}